\documentclass[lettersize,journal]{IEEEtran}
\usepackage{amsmath,amsfonts}
\usepackage{algorithmic}
\usepackage{booktabs} 
\usepackage{tabularx} 
\usepackage{algorithm}
\usepackage{array}
\usepackage{multirow}
\usepackage[caption=false,font=normalsize,labelfont=sf,textfont=sf]{subfig}
\usepackage{textcomp}
\usepackage{stfloats}
\usepackage{url}
\usepackage{verbatim}
\usepackage{graphicx}
\usepackage{cite}
\usepackage{bm}
\usepackage{amssymb}
\usepackage{xcolor}
\usepackage{soul}
\usepackage{longtable}
\begin{document}

\title{An Expectation-Maximization Algorithm-based Autoregressive Model for the Fuzzy Job Shop Scheduling Problem}
\author{Yijian Wang, Tongxian Guo, Zhaoqiang Liu
\thanks{
%Corresponding author: Zhaoqiang Liu.
The authors are with the School of Computer Science and Engineering, University of Electronic Science and Technology of China.  Emails: wangyijian@std.uestc.edu.cn; 202421081138@std.uestc.edu.cn; 
zqliu12@gmail.com. \textit{(Corresponding
author: Zhaoqiang Liu.)}}}
% The paper headers
\markboth{Journal of \LaTeX\ Class Files}%
{Shell \MakeLowercase{\textit{et al.}}: A Sample Article Using IEEEtran.cls for IEEE Journals}

\maketitle

\begin{abstract}
The fuzzy job shop scheduling problem (FJSSP) emerges as an innovative extension to the job shop scheduling problem (JSSP), incorporating a layer of uncertainty that aligns the problem more closely with the complexities of real-world manufacturing environments. This improvement increases the computational complexity of deriving the solution while improving its applicability. In the domain of deterministic scheduling, neural combinatorial optimization (NCO) has recently demonstrated remarkable efficacy. However, its application to the realm of fuzzy scheduling has been relatively unexplored. This paper aims to bridge this gap by investigating the feasibility of employing neural networks to assimilate and process fuzzy information for the resolution of FJSSP, thereby leveraging the advancements in NCO to enhance fuzzy scheduling methodologies. To achieve this, we approach the FJSSP as a generative task and introduce an expectation-maximization algorithm-based autoregressive model (EMARM) to address it. During training, our model alternates between generating scheduling schemes from given instances (E-step) and adjusting the autoregressive model weights based on these generated schemes (M-step). This novel methodology effectively navigates around the substantial hurdle of obtaining ground-truth labels, which is a prevalent issue in NCO frameworks.  In testing, the experimental results demonstrate the superior capability of EMARM in addressing the FJSSP, showcasing its effectiveness and potential for practical applications in fuzzy scheduling. 
\end{abstract}

\begin{IEEEkeywords}
Autoregressive model, expectation-maximization algorithm, fuzzy job shop scheduling problem, generative model, neural combinatorial optimization.
\end{IEEEkeywords}
\section*{Notations}
\begin{tabular}{@{}p{0.09\textwidth}@{\extracolsep{\fill}}p{0.7\textwidth}}
\textbf{Sets} \\
$\cal{J}$ & Job set \\
$\cal{M}$ & Machine set \\
$\cal{O}$ & Operation set
\end{tabular}

\begin{tabular}{@{}p{0.09\textwidth}@{\extracolsep{\fill}}p{0.35\textwidth}}
\textbf{Parameters} \\
\textit{n} & Number of jobs \\
\textit{m} & Number of machines \\
\textit{N} & Number of operations \\
\end{tabular}

\begin{tabular}{@{}p{0.09\textwidth}@{\extracolsep{\fill}}p{0.35\textwidth}}
\textbf{Variables} \\
\textit{J}$_i$ & The job to which operation $i$ belongs \\
\textit{M}$_i$ & The specific machine designated for operation $i$ \\
\textit{St}$_i$ & The first operation of the job that operation $i$ belongs \\
\textit{End}$_i$ & The last operation of the job that operation $i$ belongs \\
\end{tabular}

\begin{tabular}{@{}p{0.09\textwidth}@{\extracolsep{\fill}}p{0.35\textwidth}}
$\tilde{s}_i$ & The fuzzy start time of operation $i$ \\
$\tilde{t}_i$ & The fuzzy processing time of operation $i$ \\
\textit{o}$_{t,i}$ & The ready operation of job $j$ at step $t$ \\
\textit{I} & The instance \\
$\bm{\pi}$ & The scheduling scheme \\
\textbf{Functions} \\
$\mathrm{Defuzz}$($\cdot$) & Perform defuzzification operation on fuzzy numbers \\
$\mathrm{Quartile}$($\cdot$) & Calculate the quartiles \\
$\mathrm{FC}$($\cdot$) & Calculate the fuzzy completion time \\
$\mathrm{FMS}$($\cdot$) & Calculate the fuzzy makespan \\
\end{tabular}

\section{Introduction}
\IEEEPARstart{T}{he} job shop scheduling problem (JSSP) is a well-established combinatorial optimization problem (COP) that holds both theoretical significance and practical relevance \cite{yao2024novel,tian2023knowledge,zhang2022multitask}. The JSSP describes the processing time in the form of crisp numbers. However, in real-world manufacturing scenarios, numerous uncertain factors, such as human variability \cite{he2021multiobjective} and machine flexibility \cite{huang2024fuzzy}, often preclude the accurate specification of processing times. To overcome this limitation, the fuzzy JSSP (FJSSP) that represents processing times with fuzzy numbers has emerged and drawn extensive attention \cite{li2022self}.

Current algorithms for the FJSSP predominantly employ heuristic algorithms. Li et al. \cite{li2023bi} contributed to the field by developing a bi-population balancing multiobjective evolutionary algorithm tailored for the distributed flexible FJSSP. They introduced an innovative crossover operator and two cooperative population environmental selection mechanisms, which effectively balanced the convergence and diversity capabilities of the algorithm. Zhang et al. \cite{zhang2024decomposition} proposed a multiobjective evolutionary algorithm that integrated clustering and hierarchical estimation techniques, addressing the challenge of maintaining a balance between the diversity and convergence of nondominated solutions while ensuring overall convergence. Gao et al. \cite{gao2020solving} crafted a differential evolution algorithm with a novel selection mechanism, enhancing the solution of FJSSP. Li et al. \cite{li2020improved} enhanced an artificial immune system algorithm for the flexible FJSSP by incorporating simulated annealing, thereby boosting its exploration capabilities. Sun et al. \cite{sun2019hybrid} created an effective hybrid cooperative coevolution algorithm aimed at minimizing the fuzzy makespan of the flexible FJSSP, with a combination of particle swarm optimization and genetic algorithms significantly improving the convergence ability. Wang et al. \cite{wang2022solving} utilized fuzzy relative entropy to convert a multiobjective FJSSP into a single-objective problem and designed a hybrid adaptive differential evolution algorithm to address it. Pan et al. \cite{pan2021bi} focused on the energy-efficient flexible FJSSP and developed a bi-population evolutionary algorithm with feedback to maximize minimum agreement index and fuzzy total energy consumption and minimize fuzzy makespan.

Recognizing the potential of learning approaches, several scholars have successfully integrated them into their work, yielding promising results. Li et al. \cite{li2022learning} developed a learning-based reference vector memetic algorithm to address the multiobjective energy-efficient flexible FJSSP with type-2 fuzzy processing times. Zhao et al. \cite{zhao2022hyperheuristic} introduced a hyperheuristic algorithm with Q-learning to address the energy efficient distributed blocking flow shop scheduling problem. Deng et al. \cite{deng2023reinforcement} presented a 3-D estimation of distribution algorithm integrated with reinforcement learning, targeting the minimization of makespan and total energy consumption while maximizing delivery accuracy. They notably employed a Q-learning-based biased decoding method to adaptively adjust the evolutionary direction. Li et al. \cite{li2023co} designed a co-evolutionary algorithm leveraging deep Q-networks for the energy-aware distributed heterogeneous flexible JSSP. Focusing on the energy-efficient distributed fuzzy hybrid blocking flow shop scheduling problem, Shao et al. \cite{shao2024mql} developed a meta-Q-learning-based multiobjective search framework that automatically selected search operators to simplify the optimization process. Zhao et al. \cite{zhao2022reinforcement} tackled the energy-efficient distributed no-wait flow shop scheduling problem, prioritizing the minimization of makespan and total energy consumption, and proposed a cooperative meta-heuristic algorithm incorporating Q-learning. Pan et al. \cite{pan2022learning} addressed the flexible JSSP complicated by finite transportation resources with a learning-based multipopulation evolutionary optimization approach, featuring a reinforcement learning-based mating selection mechanism. Li et al. \cite{li2022improved} enhanced the artificial bee colony algorithm with Q-learning for the permutation flow shop scheduling problem. Lin et al. \cite{lin2022learning} proposed a learning-based grey wolf optimizer for the stochastic flexible JSSP, characterized by limited additional resources and machine-dependent setup times. Across these works, learning approaches serve a supportive role, deployed as local improvement strategies rather than as the core solvers.

The surge in deep learning has sparked significant advancements in neural combinatorial optimization (NCO), with a growing trend toward leveraging learning approaches for scheduling problems. Zhang et al. \cite{zhang2020learning} developed a deep reinforcement learning agent capable of automatically learning priority dispatch rules for the JSSP, complemented by a graph neural network to embed the states encountered during problem-solving. Kotary et al. \cite{kotary2022fast} employed a deep learning approach to provide efficient and precise approximations for the JSSP, utilizing Lagrangian duality to capture problem constraints. Corsini et al. \cite{corsini2024self} explored a self-supervised training strategy for the JSSP, where the model was trained by sampling multiple solutions and the best solution was used as a pseudo-label according to the objective of the problem. Wang et al. \cite{wang2023flexible} introduced a pioneering end-to-end learning framework that harmoniously combines self-attention models for deep feature extraction with the scalability of deep reinforcement learning for decision-making. Pan et al. \cite{pan2021deep} proposed an efficient deep reinforcement learning-based optimization algorithm to minimize the maximum completion time in the permutation flow shop scheduling. Tassel et al. \cite{tassel2023end} presented an innovative end-to-end solution that applies constraint programming and reinforcement learning to scheduling problems. Song et al. \cite{song2022flexible} contributed to the flexible JSSP with a new deep reinforcement learning method aimed at learning high-quality production dispatching rules in an end-to-end manner. Liu et al. \cite{liu2023dynamic} formulated the dynamic parallel machine scheduling as a Markov decision process to accommodate unexpected events, inspiring the development of a deep reinforcement learning method for the dynamic parallel machine scheduling. Pan et al. \cite{pan2023knowledge} proposed a knowledge-guided end-to-end optimization framework based on reinforcement learning for the permutation flow shop scheduling problem. Chen et al. \cite{chen2022deep} presented a sophisticated deep reinforcement learning framework that integrated an attention mechanism with disjunctive graph embeddings. This innovative approach leveraged a sequence-tosequence model to tackle the intricate challenges associated with the JSSP.  Du et al. \cite{du2022reinforcement} developed a deep Q-network to address the multiobjective flexible JSSP, incorporating crane transportation and setup times. Lei et al. \cite{lei2023large} proposed an end-to-end hierarchical reinforcement learning framework for the large-scale dynamic flexible JSSP, capable of near real-time operation. Li et al. \cite{li2023double} tackled the distributed heterogeneous hybrid flow shop scheduling with a dual deep Q-network-based co-evolutionary algorithm. Liu et al. \cite{liu2023integrating} integrated a deep neural network into the decomposition and coordination framework of surrogate Lagrangian relaxation to predict satisfactory solutions for the JSSP. Liu et al. \cite{liu2023dynamic2} combined graph neural networks with deep reinforcement learning for dynamic JSSP. Luo et al. \cite{luo2021real} developed a hierarchical multi-agent proximal policy optimization for managing the dynamic partial-no-wait multiobjective flexible JSSP. However, these works are based on deterministic environments, and there has been no work that employs learning approaches as the primary solver for addressing fuzzy scheduling problems. This paper aims to investigate whether neural networks can assimilate fuzzy information and apply it to solve the FJSSP, thereby integrating the advancements of NCO into fuzzy scheduling.

In this paper, we firstly model the FJSSP as a generative task and introduce an expectation-maximization (EM) algorithm-based autoregressive model (EMARM) to address it. This approach addresses the major challenge of obtaining ground-truth labels, which is a common problem in the NCO framework. Secondly, to enable the neural network to understand and process fuzzy information, we equip it with a hand-crafted prior and employ a novel method for ranking fuzzy numbers. Finally, we conduct comprehensive experiments to assess the performance of EMARM, and the results indicate that it holds an advantage over comparative algorithms, demonstrating its efficacy in solving the FJSSP.

The remainder of this paper is organized as follows. In Section II, we introduce fuzzy numbers, outline their operational properties, and define the characteristics of the FJSSP. Section III provides a detailed exposition of the proposed EMARM. Section IV presents the experimental results and analysis. Finally, Section V presents the conclusion and future work.

\section{Preliminaries}
\subsection{Fuzzy number}
Fuzzy numbers are used to represent the processing time for the fuzzy scheduling, which is described in this section.

In manufacturing environments, precise processing times are often elusive due to variables such as the diverse skill levels of workers \cite{zhu2020optimal}. While exact durations may not be predictable, experts can often draw on historical data to provide estimated durations \cite{lin2002fuzzy}. To address this unpredictability, a prevalent strategy involves estimating within confidence intervals. When certain values are more likely, opting for a fuzzy interval or number becomes an appropriate choice \cite{fortemps1997jobshop}.

Assume $S$ is a fuzzy set defined on $\mathbb{R}$, with a membership function $\mu_S: \mathbb{R} \rightarrow [0,1]$. The $\alpha$-cut of $S$ is defined as $S_\alpha = \{x \in \mathbb{R} : \mu_S(x) \geq \alpha\}, \alpha \in [0,1]$, and the support is $S_0$. A fuzzy interval is delineated by its $\alpha$-cuts being confined, and a fuzzy number $\tilde{N}$, with a compact support and a pronounced modal value, is depicted by closed intervals $\tilde{N}_\alpha = [\underline{n}_\alpha, \bar{n}_\alpha]$.

The triangular fuzzy number (TFN) \cite{deng2023reinforcement} is frequently utilized in fuzzy scheduling problems. Let $\tilde{A}$ be a TFN denoted as $\tilde{A} = (a_1, a_2, a_3)$, where $a_1$ and $a_3$ outline the range of potential values and $a_2$ signifies the modal value within this range. The membership function of $\tilde{A}$ is defined as follows:

\begin{equation}
\mu_{\tilde{A}}(x) = 
\begin{cases} 
    \frac{x-a_1}{a_2-a_1}, & \text{if } a_1 < x \leq a_2, \\
    \frac{a_3-x}{a_3-a_2}, & \text{if } a_2 < x \leq a_3, \\
    0, & \text{otherwise.}
\end{cases}
\end{equation}

Let $\tilde{A} = (a_1, a_2, a_3)$ and $\tilde{B} = (b_1, b_2, b_3)$ represent two TFNs. The operations on these TFNs are defined as follows:

\textbf{1. Addition Operation.} 
 According to \cite{nguyen2018first}, the sum of $\tilde{A}$ and $\tilde{B}$ is calculated as follows:
\begin{equation}
\widetilde{A} + \widetilde{B} = \left(a_{1} + b_{1}, a_{2} + b_{2}, a_{3} + b_{3}\right).
\end{equation}

\textbf{2. Max operation.}
According to \cite{lei2010fuzzy}, the max operation between $A$ and $B$ is defined as:
\begin{equation}
\max(\tilde{A}, \tilde{B}) = 
\begin{cases}
    \tilde{A}, & \text{if } \tilde{A} \geq \tilde{B}, \\
    \tilde{B}, & \text{otherwise}.
\end{cases}
\end{equation}

It is evident that the max operation is determined by a ranking operation. In fuzzy scheduling, a commonly used ranking method is as follows\cite{sakawa1999efficient}: Let $c_1(\tilde{A}) = \frac{a_1 + 2a_2 + a_3}{4}$, $c_2(\tilde{A}) = a_2$, and $c_3(\tilde{A}) = a_3 - a_1$. $\tilde{A} > \tilde{B}$ if and only if one of the following three conditions is met:
\begin{enumerate}
    \item $c_1(\tilde{A}) > c_1(\tilde{B})$;
    \item $c_1(\tilde{A}) = c_1(\tilde{B}) \land c_2(\tilde{A}) > c_2(\tilde{B})$;
    \item $c_1(\tilde{A}) = c_1(\tilde{B}) \land c_2(\tilde{A}) = c_2(\tilde{B}) \land c_3(\tilde{A}) > c_3(\tilde{B})$.
\end{enumerate}

However, this method can only compare the magnitude of two TFNs at a time, and comparing multiple TFNs in this way would be very time-consuming, which is not conducive to the learning process of neural networks. Another commonly used ranking method\cite{heilpern1992expected} involves defuzzifying fuzzy numbers into crisp numbers, allowing the comparison of these values to determine the relationship between the TFNs, as shown in the following formula:
\begin{equation}
\mathrm{Defuzz}(\tilde{A}) = \frac{a_1 + 2a_2 + a_3}{4}. \label{eq.4}
\end{equation}

This method can efficiently compare multiple numbers, but it only takes into account the magnitude of the expected values while neglecting the differences in boundary values. Considering both of the above points, this paper employs the following comparison method\cite{palacios2015coevolutionary}: For any TFN, we first calculate its $Z$-value using Eq. (\ref{Eq.5}):
\begin{equation}
Z(\tilde{A}) = V^\beta(\tilde{A}) + \omega S(\tilde{A}), \label{Eq.5}
\end{equation}
where $\omega$ is a weight ranging from $[0,1]$. $V^\beta(\tilde{A})$ represents the expected value, and its formula is:
\begin{equation}
V^\beta(\tilde{A}) = \beta V^{\max}(\tilde{A}) + (1-\beta) V^{\min}(\tilde{A}),
\end{equation}
where 
\begin{align}
V^{\max}(\tilde{A}) = a_2 + \int_{a_2}^{a_3} \mu_{\tilde{A}}(x) \, \mathrm{d}x, \\
V^{\min}(\tilde{A}) = a_2 - \int_{a_1}^{a_2} \mu_{\tilde{A}}(x) \, \mathrm{d}x,
\end{align}
and $\beta$ is a weight ranging from $[0,1]$. $S(\tilde{A})$ takes into account the boundary values, and its formula is:
\begin{equation}
S(\tilde{A}) = a_3 - a_1.
\end{equation}
Then, the magnitude of the $Z$-value represents the magnitude of the original number.

In addition to the well-defined operations mentioned above, this paper also addresses subtraction, division, and other operations on TFNs. The defuzzification method defined in Eq.~(\ref{eq.4}) is applied to these operations, enabling the neural network to learn the corresponding fuzzy information.
\subsection{FJSSP}
The FJSSP involves a collection of jobs, machines, and operations. Specifically, there are $n$ jobs denoted by set $\mathcal{J}$, $m$ machines represented by set $\mathcal{M}$, and $N$ operations within set $\mathcal{O}$. Each operation $i \in \mathcal{O}$ is associated with a unique job $J_i \in \mathcal{J}$, processed by a specific machine $M_i \in \mathcal{M}$, and has an uncertain processing time denoted by a TFN $\tilde{t}_i$. The operations are linked by a binary relationship $``\rightarrow"$ that forms chains for each job. If operation $i$ precedes $j$ ($i \rightarrow j$), they share the same job $J_i = J_j$, and no other operation $x$ can exist such that $i \rightarrow x$ or $x \rightarrow j$. Let $\mathcal{S}$ be the set of scheduling schemes. The objective of FJSSP is to minimize the fuzzy makespan, i.e., to find the fuzzy start time $\tilde{s}_i$ for each operation $i \in \mathcal{O}$ to minimize the following objective over all possible schemes:
\begin{equation}
\begin{aligned}
& \max_{i \in \cal{O}} \tilde{s}_i + \tilde{t}_i\\
\text{s.t.} \  &  \tilde{s}_i \geq 0, \, \forall i \in \cal{O}, \\
& \tilde{s}_j \geq \tilde{s}_i + \tilde{t}_i, \, \text{if } i \rightarrow j, i, j \in \cal{O}, \\
&\tilde{s}_j \geq \tilde{s}_i + \tilde{t}_i \lor \tilde{s}_i \geq \tilde{s}_j + \tilde{t}_j, \, \text{if } M_i = M_j, i, j \in \cal{O}. \label{eq.10}
\end{aligned}
\end{equation}
To enable the neural network to assimilate the intricate constraint information inherent in the FJSSP, this paper deviates from the conventional mixed-integer linear programming models\cite{tirkolaee2020fuzzy} typically utilized in heuristic algorithms. Instead, we adopt the disjunctive graph\cite{van1992job} to effectively model the FJSSP. This methodological choice facilitates a more nuanced representation of the scheduling constraints, enhancing the capacity of the network to learn and solve within this complex domain.

The disjunctive graph $G = (V, A, E)$ is defined as follows:
\begin{itemize}
    \item $V = \mathcal{O} \cup \{S, T\}$, where $S$ and $T$ denote the starting and ending virtual nodes, respectively, each with a processing time of zero.
    \item $A$ encompasses ordered pairs $(i, j)$ for $(i, j) \in \mathcal{O}$ such that $i \rightarrow j$, along with pairs $(S, j)$ for the first operations of all jobs and $(i, T)$ for the last operations, representing the directed connections between operations within the jobs.
    \item $E$ includes pairs $(i, j)$ where $M_i = M_j$, indicating undirected connections between operations assigned to the same machine.
\end{itemize}

For each pair of operations $i, j \in \mathcal{O}$ with $i \rightarrow j$, the second constraint in Eq. (\ref{eq.10}) is denoted as a directed edge $(i, j)$ in $A$. Similarly, for each pair of operations $i, j \in \mathcal{O}$ with $M_i = M_j$, the third constraint in Eq. (\ref{eq.10}) is denoted as an undirected edge $(i, j)$ in $E$, and the two ways of solving the disjunction correspond to the two possible orientations of the edge. Therefore, finding a scheduling scheme $\bm{\pi}$ of FJSSP is equivalent to determining the direction of each undirected edge, resulting in a directed acyclic graph. An example of a disjunctive graph for an FJSSP instance and its solution are shown in Fig. 1. When a FJSSP is modeled as a disjunctive graph, \textit{J}$_i$ (the job to which operation $i$ belongs), \textit{M}$_i$ (the specific machine designated for operation $i$), \textit{St}$_i$ (the first operation of the job that operation $i$ belongs), \textit{End}$_i$ (the last operation of the job that operation $i$ belongs), and $\tilde{t}_i$ (the fuzzy processing time of operation $i$)  are all directly obtainable. For the scheduled operation $i$, its start time $\tilde{s}_i$ is also available, and consequently, its fuzzy completion time can be calculated by $\tilde{s}_i + \tilde{t}_i$. For a scheduling scheme $\bm{\pi}$, its fuzzy makespan can also be determined.

Moreover, since the directed edges give constraints on the processing order of all the operations in each job, in decision making, we only need to determine which job (rather than which operation) needs to be processed. For example, Fig.~1(b) corresponds to a scheduling scheme (1, 3, 2, 3, 1, 2, 1, 2, 3), and the objective of the method proposed in this paper is to decide on scheduling schemes such as this one and make their fuzzy makespan as small as possible.
\begin{figure}
\begin{center}
\begin{tabular}{c}
\includegraphics[height=0.195\textwidth]{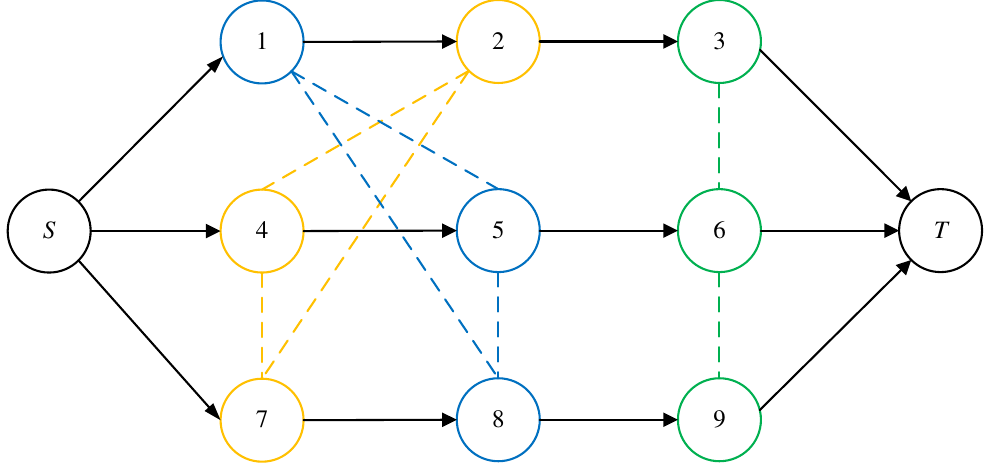} \\
{\small (a) FJSSP instance} \\
\includegraphics[height=0.195\textwidth]{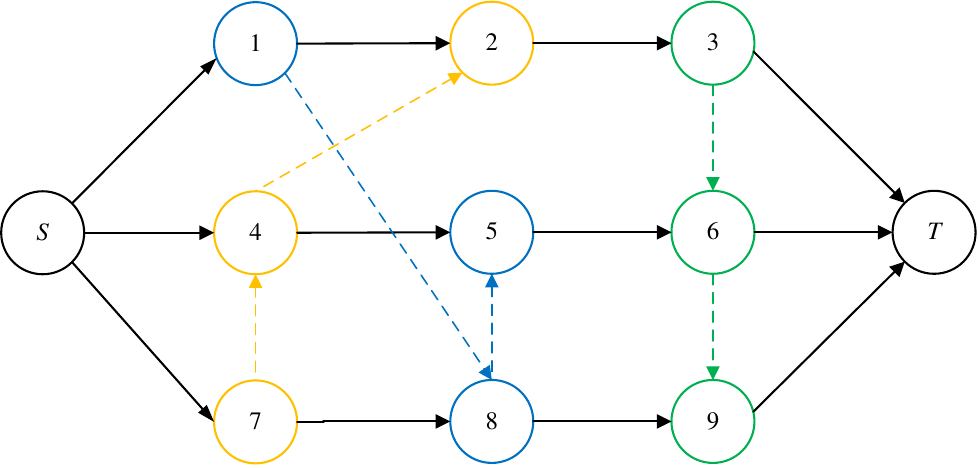} \\
{\small (b) Directed solution} \\
\end{tabular}
\caption{Disjunctive graph model. (a) illustrates a 3 × 3 FJSSP instance. The black solid lines delineate set \textit{A}, while the colored dashed lines enclose set \textit{E}. Operations with the same frame color must be executed on the same machine. (b) presents a solution for (a), where all undirected edges have been directed.}
\end{center}
\end{figure}

\section{EMARM}
In this section, we begin by formulating the FJSSP as a generative task and outlining its optimization objectives. Subsequently, we detail the computational methods for each specific component of the objective. Finally, we introduce a strategy to address the challenge posed by the absence of ground-truth solutions in NCO using the EM algorithm.
\subsection{Autoregressive model for FJSSP}
Given an training set $\left \{I_l\right \}_{l=1}^{L}$ and its scheduling scheme $\left \{\bm{\pi}_l\right \}_{l=1}^{L}$, where $\bm{\pi}_l = (\pi_{l,1},...,\pi_{l,mn})$, our objective is to model the joint probability distribution $p(I_l, \bm{\pi}_l)$ to derive the conditional probability $p(\bm{\pi}_l \mid I_l)$. By leveraging the chain rule, we can express this as:
\begin{equation}
p(\bm{\pi}_l \mid I_l) = p\left( \pi_{l,1} \mid I_l \right)\prod_{i=2}^{mn} p\left(\pi_{l,i} \mid \pi_{l,1}, \ldots, \pi_{l,i-1}, I_l\right), \label{eq.14}
\end{equation}
where $m$ denotes the number of machines, $n$ represents the number of jobs, and $mn$ is the length of $\bm{\pi}_l$. It is evident that modeling $p(\bm{\pi}_l \mid I_l)$ in this way would require exponential complexity, which is not feasible. Assuming $\pi_{l,i}$ to be i.i.d. or Markovian would make it possible to model $p(\bm{\pi}_l \mid I_l)$, but this assumption does not align with the actual situation of the FJSSP as discussed in Section II.B. To address this, we posit the existence of a neural network $p_\theta$ (parameterized by $\theta$) that approximates the conditional probability table in Eq. (\ref{eq.14}):
\begin{equation}
p_\theta(\bm{\pi}_l \mid I_l) = \prod_{i=1}^{mn} p_\theta\left(\pi_{l,i} \mid \pi_{l,1}, \ldots, \pi_{l,i-1}, I_l\right). \label{eq.15}
\end{equation}
Consequently, an autoregressive model is established, and the $\bm{\pi}_l$ is constructed by iteratively generating $\pi_{l,i}$ from $i=1$ to $mn$.

Our objective is to identify the parameter $\theta$ that maximize the log-likelihood of the optimal scheduling scheme $\left \{\bm{\pi}^{*}_l \right \}_{l=1}^{L}$ for training set $\left \{I_l\right \}_{l=1}^{L}$:
\begin{equation}
\theta = \arg \underset{\theta}{\max} \sum_{l=1}^{L} \log p_\theta\left(\bm{\pi}^{*}_l \mid I_l\right). \label{eq.16}
\end{equation}

The remainder of this section offers a comprehensive explanation of the methods for representing and calculating each element within Eqs. (\ref{eq.15}) and (\ref{eq.16}). In Sections III.B-D, we focus on a specific instance to illustrate the principles, for brevity, we omit the subscript $l$.

\subsection{Hand-crafted fuzzy prior}
In the disjunctive graph model, the data information, specifically the fuzzy processing times, is stored at the vertices representing the operations, while constraints are defined by the edges. However, the raw data for these vertices includes only a single fuzzy processing time, which is insufficient to make accurate decisions. To achieve this, we require more comprehensive information. This is because, in addition to its own numerical magnitude, its role (relative numerical magnitude) in the job and machine to which it belongs is also important. We manually design significant information as priors to be integrated into the network. For each operation $i$, the composition of its information vector $\mathbf{x}_i \in \mathbb{R}^{18}$ is as follows:
\begin{equation}
\tilde{t}_i = \left(t_1, t_2, t_3\right) \in \mathbb{R}^{3}, \label{eq.17}
\end{equation}
\begin{equation}
\mathrm{Defuzz}\left(\tilde{t}_i\right) \in \mathbb{R},
\label{eq.18}
\end{equation}
\begin{equation}
\frac{\sum_{j=\textit{St}_i}^{i} \mathrm{Defuzz}\left(\tilde{t}_j\right)}{\sum_{j=\textit{St}_i}^{\textit{End}_i} \mathrm{Defuzz}\left(\tilde{t}_j\right)} \in \mathbb{R},
\label{eq.19}
\end{equation}
\begin{equation}
\frac{\sum_{j=i+1}^{\textit{End}_i} \mathrm{Defuzz}\left(\tilde{t}_j\right)}{\sum_{j=\textit{St}_i}^{\textit{End}_i} \mathrm{Defuzz}\left(\tilde{t}_j\right)} \in \mathbb{R},
\label{eq.20}
\end{equation}
\begin{equation}
\mathrm{Defuzz}\left(\mathrm{Quartile}\left(J_i\right)\right) \in \mathbb{R}^{3},
\label{eq.21}
\end{equation}
\begin{equation}
\mathrm{Defuzz}\left(\mathrm{Quartile}\left(M_i\right)\right) \in \mathbb{R}^{3},
\label{eq.22}
\end{equation}
\begin{equation}
\mathrm{Defuzz}\left(\tilde{t}_i\right) - \mathrm{Defuzz}\left(\mathrm{Quartile}\left(J_i\right)\right) \in \mathbb{R}^{3},
\label{eq.23}
\end{equation}
\begin{equation}
\mathrm{Defuzz}\left(\tilde{t}_i\right) -\mathrm{Defuzz}\left(\mathrm{Quartile}\left(M_i\right)\right) \in \mathbb{R}^{3},
\label{eq.24}
\end{equation}
where $\mathrm{Quartile}\left(J_i\right)$ and $\mathrm{Quartile}\left(M_i\right)$ represent the quartile of the fuzzy processing time for operations belonging to $J_i$ and $M_i$, respectively. Eqs. (\ref{eq.17}) and (\ref{eq.18}) describe local information and represent the fuzzy processing time and the defuzzified processing time of $\mathcal{O}_i$, respectively. Other equations describe global information. Eqs. (\ref{eq.19}) and (\ref{eq.20}) describe how much of the job to which it belongs has been completed and how much is left after processing $\mathcal{O}_i$, respectively. Eqs. (\ref{eq.21}) and (\ref{eq.22}) describe the quartiles of fuzzy processing time for the job and machine to which $\mathcal{O}_i$ belongs, respectively. Eqs. (\ref{eq.23}) and (\ref{eq.24}) describe the difference of the defuzzified fuzzy processing time of $\mathcal{O}_i$ and Eqs. (\ref{eq.21}) and (\ref{eq.22}), respectively.

\subsection{Encoder}
A two-layer graph attention network (GAT)\cite{brody2021attentive} is employed to extract and learn valuable information from the disjunctive graph. This network transforms 18-dimensional $\mathbf{x}_i$ into a $h$-dimensional $\mathbf{e}_i$. The primary advantage of $\mathbf{e}_i$ over $\mathbf{x}_i$ lies in its ability to complement the relationship between edges. To preserve the vertex information without dilution, the output at each layer of the GAT is concatenated with the original feature vector $\mathbf{x}_i$. The formulation of encoder is detailed as follows:
\begin{equation}
\mathbf{e}_i = \left[ \mathbf{x}_i \| \sigma\left( \operatorname{GAT}_2\left( \left[ \mathbf{x}_i \| \sigma\left( \operatorname{GAT}_1\left(\mathbf{x}_i, D \right) \right) \right], D \right) \right) \right],
\end{equation}
where $``\|"$ is the concatenation operation, $\sigma$ is the ReLU activation function, $D = A \cup E$ is the set of edges for instance $I$, and $\mathbf{e}_i$ describes the instance $I$ in Eq. (\ref{eq.15}).

\subsection{Decoder}
The decoder is composed of two parts: A state network and a decision network. The former is responsible for updating state, and the latter is responsible for making decisions based on the information provided by the encoder and the state network. The two networks are described below.

\textbf{1. The state network.} EMARM makes decisions step by step, and each decision impacts the future. Therefore, after each decision, it is necessary to update the state and convey it to the EMARM to enhance the accuracy of subsequent decisions.

First, we generate an 11-dimensional context vector $\mathbf{c}_i$ ($i = 1, 2, \ldots, n$) for each job to describe its information. Assuming that $o_{t,i}$ denote the ready operation of job $\mathcal{J}_i$ at step $t$ and its predecessor is $o_{t,i}-1$. The context vector $\mathbf{c}_i \in \mathbb{R}^{11}$ contains the following entries:
\begin{equation}
\mathrm{Defuzz}\left(\mathrm{FC}(o_{t, i}-1)\right) - \mathrm{Defuzz}\left(\mathrm{FC}(M_{o_{t, i}})\right) \in \mathbb{R},
\label{eq.26}
\end{equation}
\begin{equation}
\frac{\mathrm{Defuzz}\left(\mathrm{FC}(o_{t, i}-1)\right)}{\mathrm{Defuzz}\left(\max_{i=1,\ldots, n} \mathrm{FC}(\mathcal{J}_{i})\right)} \in \mathbb{R},
\label{eq.27}
\end{equation}
\begin{equation}
\mathrm{Defuzz}\left(\mathrm{FC}(o_{t, i}-1)\right) - \frac{\mathrm{Defuzz}\left(\sum_{i=1}^{n} \mathrm{FC}(\mathcal{J}_{i})\right)}{n} \in \mathbb{R},
\label{eq.28}
\end{equation}
\begin{equation}
\mathrm{Defuzz}\left(\mathrm{FC}(o_{t, i}-1)\right) - \mathrm{Defuzz}\left(\mathrm{Quartile}(\mathcal{J})\right) \in \mathbb{R}^{3},
\label{eq.29}
\end{equation}
\begin{equation}
\frac{\mathrm{Defuzz}\left(\mathrm{FC}(M_{o_{t,i}})\right)}{\mathrm{Defuzz}\left(\max_{i=1,\ldots, m} \mathrm{FC}(\mathcal{M}_{i})\right)} \in \mathbb{R},
\label{eq.30}
\end{equation}
\begin{equation}
\mathrm{Defuzz}\left(\mathrm{FC}(M_{o_{t,i}})\right) - \frac{\mathrm{Defuzz}\left(\sum_{i=1}^{m} \mathrm{FC}(\mathcal{M}_{i})\right)}{m} \in \mathbb{R},
\label{eq.31}
\end{equation}
\begin{equation}
\mathrm{Defuzz}\left(\mathrm{FC}(M_{o_{t,i}})\right) - \mathrm{Defuzz}\left(\mathrm{Quartile}(\mathcal{M})\right) \in \mathbb{R}^{3},
\label{eq.32}
\end{equation}
where $\mathrm{FC}(\cdot)$ calculates the fuzzy completion time. Eq. (\ref{eq.26}) describes the relationship between two factors that affect $o_{t,i}$ processing: Whether the predecessor operation is complete and whether the required machine is idle. Eq. (\ref{eq.27}) measures how close the current fuzzy completion time of $J_{o_{t, i}}$ is to the current fuzzy makespan. Eq. (\ref{eq.28}) measures how early or late the fuzzy completion time of $J_{o_{t, i}}$ is compared to the average fuzzy completion time of all jobs in current state. Eq.~(\ref{eq.29}) describes the relative fuzzy completion time of $J_{o_{t, i}}$ with respect to other jobs in current state. Eq. (\ref{eq.30}) measures how close the current fuzzy completion time of $M_{o_{t, i}}$ is to the current fuzzy makespan. Eq.~(\ref{eq.31}) measures how early or late the fuzzy completion time of $M_{o_{t, i}}$ is compared to the average completion time of all jobs in current state. Eq. (\ref{eq.32}) describes the relative fuzzy completion time of $M_{o_{t, i}}$ with respect to other machines in current state.

Second, these context vectors are further integrated by Eq.~(\ref{eq.33}) to derive the state vectors $\mathbf{s}_i \in \mathbb{R}^d \left( i = 1, 2, \ldots, n \right)$. These state vectors assist the decision network in making informed decisions, where $d$ is a hyper-parameter.
\begin{equation}
\mathbf{s}_i = \sigma\left( \left[ \mathbf{c}_i \mathbf{W}_1 + \underset{j=1,\ldots, n}{\operatorname{MHA}}\left( \mathbf{c}_j \mathbf{W}_1 \right) \right] \mathbf{W}_2 \right), \label{eq.33}
\end{equation}
where $\mathbf{W}_1$ and $\mathbf{W}_2$ are learnable parameter matrices, $\operatorname{MHA}$ denotes the multi-head attention layer\cite{vaswani2017attention}, and $\mathbf{s}_i$ represents the impact of the scheduled operations on subsequent decision-making, corresponding to the $\pi_1, \ldots, \pi_{i-1}$ in Eq.~(\ref{eq.15}).

\textbf{2. Decision network.}
This network combines the $\mathbf{e}_{o_{t, i}}$ generated by the encoder, which contain global information about the FJSSP, with the $\mathbf{s}_i$ generated by the state network, which contain local state information, to generate the probability of choosing a job for the step $t$. The details are as follows:
\begin{equation}
z_i = \operatorname{FNN}\left( \left[\mathbf{e}_{o_{t, i}}\| \mathbf{s}_i \right] \right),
\end{equation}
where FNN denotes the feedforward neural network\cite{glorot2010understanding}. Subsequently, these $ z_i \in \mathbb{R} $ are transformed into probabilities with a Softmax function: $ p_i = e^{z_i} / \sum_{j=1}^{n} e^{z_j} $, where $p_i$ corresponds to $p_\theta\left(\pi_{l,i} \mid \pi_{l,1}, \ldots, \pi_{l,i-1}, I_l\right)$ in Eq. (\ref{eq.15}). Finally, the job to be processed at step $t$ is determined through the sample method described below.

On the one hand, we randomly select a job $i$ with a probability proportional to $p_i$, which is produced at step $t$ by the decoder. On the other hand, to avoid selecting jobs that have already been completed, we employ a masking operation to set the probability of selecting them in subsequent steps to zero.
    	\newcounter{TempEqCnt}
	\setcounter{TempEqCnt}{5}
	\setcounter{equation}{31}
	
	\begin{figure*}[b]
		\hrulefill
		\begin{align}
			\log p_\theta(\mathbf{y}) = \log \int p_\theta(\mathbf{y},\mathbf{x}) \, \mathrm{d}\mathbf{x} &\geq \int p_\theta(\mathbf{x} \mid \mathbf{y}) \log \frac{p_\theta(\mathbf{y},\mathbf{x})}{p_\theta(\mathbf{x} \mid \mathbf{y})} \, \mathrm{d}\mathbf{x} \notag  \\
&= \int p_\theta(\mathbf{x} \mid \mathbf{y}) \log p_\theta(\mathbf{y},\mathbf{x}) \, \mathrm{d}\mathbf{x} - \int p_\theta(\mathbf{x} \mid \mathbf{y}) \log p_\theta(\mathbf{x} \mid \mathbf{y}) \, \mathrm{d}\mathbf{x} \\
&= \mathbb{E}_{\mathbf{x} \sim p_\theta(\mathbf{x} \mid \mathbf{y})} \left[ \log p_\theta(\mathbf{y} \mid \mathbf{x}) + \log p_\theta(\mathbf{x}) - \log p_\theta(\mathbf{x} \mid \mathbf{y}) \right] \triangleq \mathcal{L}(\theta). \notag 
		\end{align}
	\end{figure*}
	\setcounter{equation}{\value{TempEqCnt}}

\begin{algorithm}[t]
\renewcommand{\algorithmicrequire}{\textbf{Input:}}
\renewcommand{\algorithmicensure}{\textbf{Output:}}
\caption{The EM algorithm-based training strategy}
\begin{algorithmic}[1]
\REQUIRE Training set $\left \{{I}_l\right \}_{l=1}^{L}$, number of epochs $T$, number of sampling times $K$
\ENSURE The parameters $\theta$ of the EMARM 
\STATE Randomly initialize the parameter $\phi$
\FOR{$t=1$ \textbf{to} $T$}
    \STATE // E-step
    \FOR{$l=1$ \textbf{to} $L$}
    \STATE Calculate $p_\phi(\bm{\pi}_l \mid {I}_l)$ according to Eq. (\ref{eq.15})
    \STATE Sample $\left \{\bm{\pi}_l^1, \ldots, \bm{\pi}_l^K \right \} \sim  p_\phi(\bm{\pi}_l \mid {I}_l)$
    \STATE $\hat{\bm{\pi}}_{l} = \arg\min_{\bm{\pi} \in \left\{ \bm{\pi}_{l}^{1}, \ldots, \bm{\pi}_{l}^{K} \right\}}  \mathrm{FMS}(\bm{\pi})$   
    \ENDFOR
    \STATE // M-step
    \STATE $\theta = \arg \underset{\phi}{\max} \sum_{l=1}^{L} \log p_\phi\left(\hat{\bm{\pi}}_{l} \mid {I}_l\right)$
     \STATE $\phi = \theta$
\ENDFOR
\end{algorithmic}
\end{algorithm}

\subsection{Training strategy based on EM algorithm}
To address the challenge of determining the parameter $\theta$ of EMARM according to Eq. (\ref{eq.16}), we encounter a chicken-and-egg dilemma: Knowing the optimal scheduling scheme $\bm{\pi}^*$ is essential for identifying $\theta$, yet discovering $\bm{\pi}^*$ itself relies on having a well-tuned $\theta$.

The EM algorithm\cite{dempster1977maximum} provides a solution to this dilemma. It is an iterative method designed for estimating parameters in statistical models that include latent variables. When the true values of these latent variables remain unknown, traditional maximum likelihood estimation is not feasible, and thus, the EM algorithm maximizes a lower bound of the log-likelihood function. This bound is established using Jensen's inequality, as illustrated in Eq. (32), where $\mathbf{x}, \mathbf{y},$ and $\theta$ denote the latent variables, observations, and model parameters, respectively. The algorithm proceeds through two main steps:

\begin{itemize}
    \item \textbf{Expectation step (E-step):} Sample latent variables from the current conditional distribution estimate $\mathbf{x}\sim~p_{\theta}(\mathbf{x} \mid~\mathbf{y})$, and compute the expected log-likelihood lower bound $\mathcal{L}(\theta)$.
    \item \textbf{Maximization step (M-step):} Maximize $\mathbb{E}_{\mathbf{x} \sim p_\theta(\mathbf{x} \mid \mathbf{y})} \left[ \log p_\theta(\mathbf{x}) \right]$ to update parameter $\theta$.
\end{itemize}

This iterative process enables the EM algorithm to converge towards a local maximum of the log-likelihood of the observed data, rendering it a robust technique for estimation problems with latent variables. Given the NP-hard nature of the FJSSP\cite{gao2020solving}, the optimal scheduling scheme $\bm{\pi}^*$ shares the same unobservable characteristic as the latent variable. This similarity motivates us to tackle the problem using the EM algorithm, the pseudo-code for which is presented in Algorithm 1, where $\bm{\pi}_l$ is a scheduling scheme of instance $I_l$, $\mathrm{FMS}(\cdot)$ calculates the fuzzy makespan. Furthermore, in the specific implementation, the expected solution for each instance may be obtained by sampling multiple candidate scheduling schemes and selecting the best, thereby improving the performance of EMARM in the early stages.

\section{Numerical Result and Comparisons}
In this section, numerous experiments are performed to validate the effectiveness of the EMARM. The EMARM is developed using Python 3.9 and PyTorch 1.3.1, and is executed on an Ubuntu 22.04 PC. The hardware setup includes an Intel Platinum 8358P processor and an NVIDIA GeForce RTX 4090 with 24GB of memory.
\subsection{Dataset and test instances}
We randomly generate 30000 instances as the training set by following \cite{li2021many}. The size ($m \times n$) of the training set includes $10 \times 10$, $15 \times 10$, $15 \times 15$, $20 \times 10$, $20 \times 15$, and $20 \times 20$ with each size represented by 5000 instances. The validation set is generated in the same way as the training set, except that the number of instances for each size is 100. In order to test the performance of EMARM comprehensively, we select 9 benchmarks from \cite{abdullah2014fuzzy}, and generate 2 benchmarks (20 × 20, 30 × 20) following the method in \cite{abdullah2014fuzzy},  with each benchmark consisting of 4 instances. It is worth noting that the benchmarks are not exactly the same size as those in the training set, which helps to demonstrate the generalization ability of EMARM.

\subsection{Architecture, hyperparameters, and training}
In the encoder, we utilize two GATs, each equipped with 3 attention heads and a Leaky-ReLU slope of 0.15. In GAT$_1$, the size of each head is set to 64 and their outputs are concatenated. In GAT$_2$, the size of each head is set to 128 and their outputs are averaged. Therefore, $h = 16 + 128 = 144$, and $\mathbf{e}_i \in \mathbb{R}^{144}$.

In the state network, the size of each head in MHA layer is set to 64 and their outputs are concatenated. This results in parameter matrices $\mathbf{W}_1 \in \mathbb{R}^{11 \times 192}$ and $\mathbf{W}_2 \in \mathbb{R}^{192 \times 128}$. Therefore, $d = 128$ and $\mathbf{s}_i \in \mathbb{R}^{128}$. In the decision network, the FNN is implemented by a dense layer of 128 neurons, with a final layer of 1 neuron and Leaky-ReLU (slope = 0.15) activation function. We utilize the Adam optimizer\cite{kingma2014adam} for training EMARM. The training parameters are set as follows: The number of epochs $T$ is set to 30, the learning rate is 0.0002, and the batch size is configured at 16. During the training stage, sampling times $K$ is 256 and this number is increased to 512 during the testing stage. According to \cite{sun2019hybrid}, the $\beta$ and $\omega$ in Eq. (\ref{Eq.5}) are set to 0.5 and 0.4, respectively.

\subsection{Comparisons to other algorithms}
In this section, NSODE \cite{gao2020solving}, HADE \cite{wang2022solving}, genetic algorithm (GA) \cite{mitchell1998introduction}, and particle swarm optimization (PSO) \cite{kennedy1995particle} are selected for comparison experiments. NSODE and HADE are recent algorithms proposed in 2020 and 2022 respectively, while GA and PSO are well-established and classic evolutionary algorithms. All these algorithms are implemented in C++. The population size for all algorithms is set to 100, with each running through 100 iterations to ensure a fair comparison. Additional parameters specific to each algorithm are detailed in Table I.

NSODE is designed to address optimization objectives within the FJSSP, considering both fuzzy execution and completion times. HADE employs fuzzy relative entropy to tackle a multi-objective FJSSP, balancing fuzzy completion times with total fuzzy energy consumption. GA and PSO are representative of evolutionary computation and swarm intelligence, respectively.

\begin{table}[t]
\centering
\caption{Parameter Settings}
\tabcolsep9pt
\begin{tabular}{cc} % 使用 c 表示居中对齐，| 表示列之间有竖线
\toprule
\textbf{Algorithm} & \textbf{Parameter settings} \\
\midrule
NSODE & mutation factor: 0.5, crossover factor: 0.5 \\
HADE & mutation factor: 0.5, crossover factor: 0.5 \\
PSO & global learning factor: 1.2, local learning factor: 1.2 \\
GA & crossover probability: 0.7, mutation probability: 0.1 \\
\bottomrule
\end{tabular}
\end{table}

\begin{table}[t]
\centering
\caption{Running Time (in seconds) for EMARM, \\NSODE, HADE, PSO, and GA}
\tabcolsep9pt
\begin{tabular}{cccccc} % 定义五列，每列居中对齐
\toprule
\textbf{Size} & \textbf{EMARM} & \textbf{NSODE} & \textbf{HADE} & \textbf{PSO} & \textbf{GA} \\
\midrule
6$\times$6 & \textbf{0.118} & 0.542 & 0.801 & 1.845 & 0.935 \\
10$\times$5 & \textbf{0.133} & 0.681 & 0.915 & 2.294 & 1.150 \\
10$\times$10 & \textbf{0.226} & 1.241 & 1.544 & 4.088 & 2.120 \\
15$\times$5 & \textbf{0.275} & 0.953 & 1.137 & 3.196 & 1.639 \\
15$\times$10 & \textbf{0.475} & 1.821 & 2.489 & 6.301 & 2.984 \\
15$\times$15 & \textbf{0.435} & 2.950 & 4.471 & 10.220 & 4.578 \\
20$\times$5 & \textbf{0.406} & 1.226 & 1.506 & 4.099 & 2.063 \\
20$\times$10 & \textbf{0.407} & 2.534 & 3.700 & 8.824 & 3.970 \\
20$\times$15 & \textbf{0.605} & 4.234 & 6.740 & 14.870 & 6.601 \\
20$\times$20 & \textbf{0.768} & 6.196 & 10.190 & 22.430 & 9.477 \\
30$\times$20 & \textbf{1.225} & 18.471 & 19.490 & 41.520 & 18.940 \\
\bottomrule
\end{tabular}
\end{table}

\begin{table*}[htbp]
\centering
\caption{Fuzzy Makespan for EMARM, NSODE, HADE, PSO, and GA}
\tabcolsep12pt
\renewcommand{\arraystretch}{1.2}
\begin{tabular}{ccccccc} % 定义五列，每列居中对齐
\toprule
\textbf{Size} & \textbf{Instance} & \textbf{EMARM} & \textbf{NSODE} & \textbf{HADE} & \textbf{PSO} & \textbf{GA} \\
\midrule
\multirow{4}{*}{6$\times$6} & S6.1 & \textbf{(57,83,108)} & (63,88,107) & (61,87,110) & (97,139,176) & (62,87,107) \\

&S6.2 & \textbf{(51,70,86)} & (60,87,110) & (60,86,107) & (94,137,168) & (62,89,111) \\

& S6.3 & \textbf{(50,65,84)} & (60,79,103) & (53,71,93) & (107,151,194) & (60,78,105) \\

& S6.4 & \textbf{(27,36,45)} & (32,43,54) & (31,41,51) & (49,69,89) & (31,42,59) \\

\multirow{4}{*}{10$\times$10} & S10.1 & \textbf{(99,134,171)} & (146,202,255)	& (131,182,227)	& (236,319,399)	&(143,202,249) \\

&S10.2 & \textbf{(94,134,168)}&	(138,185,224)&	(121,166,209)	&(209,279,354)&	(145,191,250) \\

& S10.3 &\textbf{(85,116,143)}	&(139,183,239)	&(117,159,198)	&(140,195,250)	&(146,200,264) \\

& S10.4 & \textbf{(30,48,63)}	&(40,70,96)&	(39,68,94)	&(56,102,139)&	(45,79,107)\\

\multirow{4}{*}{15$\times$5} & La6 & \textbf{(667,926,1186)} &	(815,1097,1421)	&(792,1070,1369)	&(1637,2211,2822)	&(1030,1358,1722) \\

&La7 & \textbf{(821,890,1003)}	&(1137,1229,1387)	&(1064,1151,1278)&	(1604,1743,1946)&	(1240,1345,1499) \\

& La8 & \textbf{(796,863,952)}	&(1013,1144,1245) 	&(967,1103,1181)	&(1917,2174,2354)	&(1213,1366,1495) \\

& La9 & \textbf{(856,951,1014)}	&(1222,1359,1483) 	&(1101,1188,1287)	&(1783,1937,2098)&(1317,1459,1602)\\

\multirow{4}{*}{15$\times$10} & La21 &\textbf{(1051,1147,1243)}	&(1916,2052,2188)&	(1441,1569,1697)	&(2194,2350,2506)	&(2202,2342,2482) \\

&La22 &\textbf{(880,982,1125)}	&(1492,1672,1918) 	&(1366,1532,1772)	&(2057,2310,2634)&	(1572,1738,1997) \\

& Lei1 &\textbf{(144,213,279)}	&(222,320,414)	&(221,312,410)&	(309,450,580)	&(268,387,507) \\

& Lei2 & \textbf{(130,177,237)}	&(219,310,405)	&(182,253,335)	&(251,350,456)	&(225,322,409) \\

\multirow{4}{*}{15$\times$15} & La36 & \textbf{(1174,1326,1504)}& 	(2176,2418,2681)	& (1765,1958,2173)	& (2658,2938,3309)	& (2308,2556,2821) \\

&La37 & \textbf{(1354,1502,1677)}& 	(2441,2708,2993)	& (1919,2125,2329)	& (2594,2887,3247)	& (2402,2681,2988) \\

& La38 & \textbf{(1205,1297,1389)}	& (2195,2347,2499)	& (1900,2061,2222)	& (2249,2407,2565)	& (2295,2453,2611) \\

& La39 & \textbf{(1195,1322,1463)} &	(2220,2454,2741)	 &(1984,2177,2413)	 &(2324,2613,2900) &	(2320,2602,2889) \\

\multirow{4}{*}{20$\times$5} & La11 &\textbf{(1186,1315,1430)}&	(1751,1868,1985) 	&(1539,1636,1733)	&(2489,2642,2795)	&(1819,1931,2043) \\

&La12 & \textbf{(1154,1241,1328)}&	(1216,1381,1493) &	(1187,1339,1433)&	(2139,2391,2561)	&(1519,1708,1826) \\

& La13 & \textbf{(1190,1364,1501)}	&(1496,1666,1818) 	&(1398,1575,1731)	&(2533,2816,3071)&	(1706,1931,2106) \\

& FT20 & \textbf{(1129,1223,1141)}	 & (1742,1844,1944) 	 & (1478,1557,1633)	 & (2604,2728,2844) & 	(1894,1987,2077) \\

\multirow{4}{*}{20$\times$10} & La27 & \textbf{(1254,1325,1396)}	&(2555,2757,2959)	&(2044,2178,2312)	&(3159,3382,3605)	&(2592,2777,2962) \\

&La28 & \textbf{(1186,1315,1430)}	&(2356,2662,2924)	&(1806,2012,2215)&	(2607,2887,3183)&	(2461,2770,3018) \\

&La29 &\textbf{(1154,1241,1328)}	&(2305,2478,2651)	&(1884,2003,2122)	&(3047,3248,3449)	&(2556,2747,2938) \\

& La30 & \textbf{(1190,1364,1501)}	&(2419,2697,2922)	&(1848,2090,2289)&	(3019,3344,3636)&	(2782,3136,3470) \\

\multirow{4}{*}{20$\times$15} & ABZ7 & \textbf{(673,713,753)} &	(1366,1462,1558)	 &(1075,1150,1225)	 &(1909,2040,2171)	 &(1415,1496,1577) \\

&ABZ8 & \textbf{(696,721,746)}	&(1453,1518,1583)	&(1076,1153,1230)&	(1652,1751,1850)	&(1470,1562,1654) \\

& ABZ9 & \textbf{(721,764,807)}&	(1430,1514,1598)	&(1097,1180,1263)	&(1582,1669,1756)	&(1444,1530,1616) \\

& DMU1 &\textbf{(2536,2858,3094)}	&(5428,6038,6585)	&(4290,4767,5250)	&(5382,6066,6684)	&(5471,6150,6681) \\

\multirow{4}{*}{20$\times$20} &Ta21 & \textbf{(1665,1821,1971)}	&(3352,3613,3874)&	(2652,2811,2970)	&(3480,3715,3950)	&(3516,3747,3978) \\

&Ta22 & \textbf{(1638,1764,1890)}	&(3387,3651,3915)	&(2713,2936,3159)&	(3411,3704,3997)	&(3281,3566,3851) \\

&Ta23 & \textbf{(1630,1731,1832)}&	(3452,3669,3886)	&(2617,2796,2975)	&(3482,3730,3978)&	(3563,3799,4035) \\

&Ta24 & \textbf{(1648,1797,1946)}	&(3485,3755,4025)	&(2765,2997,3229)&	(3420,3694,3968)&	(3536,3816,4096)\\

\multirow{4}{*}{30$\times$20} & Ta41 & \textbf{(2137,2265,2393)}	 & (5807,6247,6687) 	 & (3801,4071,4341)	 & (4796,5159,5522)	 & (5884,6319,6754) \\

&Ta42 & \textbf{(2015,2184,2353)}	&(5205,5580,5955)	&(3679,3939,4199)	&(5264,5622,5980)	&(5445,5866,6287) \\

& Ta43 & \textbf{(1958,2115,2272)}	& (5417,5788,6159)	& (3258,3487,3716)	& (5075,5388,5701)& 	(5538,5932,6326) \\

& Ta44 & \textbf{(2073,2217,2361)}&	(5073,5431,5789)	&(3814,4074,4334)	&(5340,5728,6116)	&(5395,5763,6131)\\
\bottomrule
\end{tabular}
\end{table*}

We conduct comparisons on two aspects: Solution quality and running time. Given that evolutionary algorithms are stochastic in nature, we take the average of 30 experimental runs for the comparison. Since EMARM can be used continuously without the need for retraining after it has been trained, we only accounted for its testing time in terms of the time metric. Although EMARM and comparative algorithms are written in different languages, for fuzzy makeapsn, the difference in compilation languages does not significantly affect the quality of the solution; for running time, C++ is usually faster than Python for the same task, so this is a better indication of the superiority of our method.

The results of the running time comparison experiment are presented in Table II. Due to the similarity in running time for instances of the same size, we have opted not to list individual times for the four instances but instead provide the average running time as the experimental result. As evidenced by Table II, EMARM demonstrates a significantly reduced  running time when compared to the other algorithms. This is attributed to the fact that EMARM is a data-driven direct solver capable of solving problems through numerical operations after parameter learning.  In contrast, evolutionary algorithms rely on an iterative approach grounded in population evolution, necessitating progressive refinement of solutions across generations. Generally, iterative methods tend to require more time than direct solution methods.

Comparison experiments for fuzzy makespan are presented by Table III, which shows that EMARM outperforms all comparative algorithms in terms of solution quality. This confirms the soundness of modeling FJSSP as a generative task and the potential of learning-based approaches to solve this type of COPs.

\section{Conclusion and Future Work}
In this paper, we investigate the FJSSP. This problem extends the JSSP by employing fuzzy numbers to characterize uncertainties in the production process. While this approach is more closely aligned with real-world manufacturing scenarios, it also increases the complexity of finding solutions. NCO is experiencing rapid development and is achieving many remarkable results. However, no researchers have attempted to address the FJSSP using learning-based methods.
We model the FJSSP as a generative task and propose the EMARM to solve it. Firstly, to enable neural networks to capture fuzzy information, we design hand-crafted fuzzy priors. Secondly, we employ the EM algorithm to overcome the challenge of obtaining ground-truth labels in NCO. Lastly, a large number of experiments prove the effectiveness of the method we propose in this paper.

In the future, on one hand, we intend to investigate the potential of alternative generative models in tackling the FJSSP. This includes models such as diffusion models \cite{ho2020denoising}, generative adversarial networks \cite{goodfellow2014generative}, and flow-based models \cite{dinh2014nice}. On the other hand, we also plan to apply the model presented in this paper to solve other scheduling problems, such as the permutation flow shop scheduling problem \cite{brammer2022permutation}, hybrid flow shop scheduling problem \cite{wang2024tree}, and parallel machine scheduling problem\cite{wu2023biobjective}.

\bibliography{ref}
\bibliographystyle{IEEEtran}

\end{document}